\documentclass[conference]{IEEEtran}
\IEEEoverridecommandlockouts
\usepackage{cite}
\usepackage{amsmath,amssymb,amsfonts}
\usepackage{algorithmic}
\usepackage{graphicx}
\usepackage{textcomp}
\usepackage{xcolor}
\usepackage{hyperref}       
\usepackage{booktabs}       
\usepackage{subcaption}

\begin{document}

\title{Reinforcement Learning Based Power Grid Day-Ahead Planning and AI-Assisted Control}

\author{
\IEEEauthorblockN{Anton R. Fuxjäger$^*\text{,}$ Kristian Kozak$^*\text{,}$ Matthias Dorfer$^*\text{,}$ Patrick M. Blies, Marcel Wasserer}\thanks{$^*$Authors contributed equally.}
\IEEEauthorblockA{\textit{enliteAI} \\
Vienna, A-1010 \\
\texttt{m.wasserer@enlite.ai}}
}

\maketitle

\begin{abstract}

The ongoing transition to renewable energy is increasing the share of fluctuating power sources like wind and solar, raising power grid volatility and making grid operation increasingly complex and costly. 

In our prior work, we have introduced a congestion management approach consisting of a redispatching optimizer combined with a machine learning-based topology optimization agent. Compared to a typical redispatching-only agent, it was able to keep a simulated grid in operation longer while at the same time reducing operational cost.
Our approach
also ranked 1st in the L2RPN 2022 competition initiated by RTE, Europe’s largest grid operator.
The aim of this paper is to bring this promising technology closer to the real world of power grid operation.
We deploy RL-based agents in two settings resembling established workflows, AI-assisted day-ahead planning and real-time control, in an attempt to show the benefits and caveats of this new technology.
We then analyse congestion, redispatching and switching profiles, and elementary sensitivity analysis providing a glimpse of operation robustness.
While there is still a long way to a real control room, we believe that this paper and the associated prototypes help to narrow the gap and pave the way for a safe deployment of RL agents in tomorrow’s power grids.

\end{abstract}


\section{Introduction}
\label{sec:introduction}
The energy sector is facing a rapid transition towards clean, renewable energy.
However, the integration of a growing share of volatile renewable sources into a grid that was originally designed for steady power flows will lead to novel control complexities. Reinforcing power grids with safety margins large enough to allow static operation would require unjustifiable grid investments. Instead, operators will have to push the grid closer to its limit. This trend is already having an impact on the grid, as grid congestions are becoming more frequent. Congestion mitigation techniques such as redispatching are costly, and often lead to higher emissions from fossil generators, being in direct opposition to decarbonizing the power grid
(please refer to \cite{marot2022perspectives} and \cite{viebahn2022potential} for additional background information).
In this work we address these problems and propose how to utilize a topology-based RL-agent as a complementary non-costly congestion management measure.

\textbf{Machine learning and power grids.}
Advances in Machine Learning, in particular its branch of Reinforcement Learning (RL), have demonstrated impressive capabilities in a variety of domains. RL-based agents have shown superhuman performance in games that require extensive planning such as Chess or Go~\cite{silver2017alphazero, schrittwieser2020muzero}, mastered visually complex domains like Atari games~\cite{schrittwieser2020muzero}, made industrial cooling systems more efficient~\cite{luo2022cooling}, and controlled an aircraft~\cite{osipychev2022airtraffic}.
Recently, researchers achieved similar success
in the context of power grids
by developing first agents surpassing human performance
in the task of topology-based grid optimization~\cite{marot2021retrospective}.

RL agents are trained through repeated trial and error interactions with a simulation environment. Power grid operators already routinely utilize simulators based on the well-established load flow calculation~\cite{stott1974loadflow} to simulate possible future states of the power grid. The existence of fast and reliable simulators~\cite{marot2019l2rpn} paves the way for harnessing the power of reinforcement learning in the power grid domain.

\textbf{Our contributions.}
In prior work~\cite{dorfer2022topo}, we have 
introduced a congestion management approach consisting of a redispatching optimizer combined with an RL-based topology optimization agent that was able to keep a simulated grid in operation longer than a redispatching-only approach, while at the same time reducing redispatching costs.
This 
approach has been evaluated on the Grid2Op framework~\cite{donnot2020grid2op} developed by RTE, Europe's largest grid operator, and ranked 1st%
\footnote{Our submission: \url{https://github.com/enlite-ai/maze-l2rpn-2022-submission}}
in last year's Learning to Run a Power Network (L2RPN) competition%
\footnote{Learning to Run a~Power Network, \cite{marot2019l2rpn, marot2021retrospective, marot2021l2rpntrust}, \url{https://l2rpn.chalearn.org/}}.

In this work, we present and evaluate approaches how to integrate this agent into existing grid operation workflows. This enables operators to utilize suggestions from RL-powered agents while keeping operation procedures and human control in place.
In particular, we propose and demonstrate:

\begin{itemize}
    \item An integration of our RL-based agent into the day-ahead planning workflow, comparing its actions to redispatching-only approaches in terms of congestion profiles and operational costs.
    \item An AI-assisted workflow enabling grid operators to leverage remedial action suggestions of our congestion management agent in real time\footnote{A prototype application utilizing a Grid2Op environment can be found at: \url{http://grid-demo.enlite.ai/}}
    \item A trade-off parameter to balance redispatching and topology actions, allowing 
    the operator to choose between operational plans that perform more switching operations and less redispatching or vice versa.
\end{itemize}

The aim of this paper is to narrow the gap between promising RL-based approaches and the tools that grid operators utilize every day, hopefully enabling grid operators to benefit from reduced complexity, operational costs and fossil emissions while maintaining control, safety, and reliability of the grid.
Prior work already motivates and emphasizes the need for assisting tools helping to operate the future power grid~\cite{prostejovsky2019future,marott2022towards,viebahn2022potential}.
They also suggest design principles for how to structure and build operational decision support systems along with both, technical as well as psychological foundations to underline their suggestions.
In Section~\ref{sec:day_ahead_planning} and \ref{sec:ai_assisted_control} we pick up this work and propose two concrete instances for such assistance tools
for the tasks \textit{day-ahead congestion management planning} and \textit{real time remedial action recommendation}.

\section{Background}
In this section, we give a short introduction to reinforcement learning, summarize the simulation environment as well as the method behind our congestion management agent. 
\label{sec:background}
\subsection{Reinforcement Learning}

In RL, an agent is usually trained by interacting with a provided simulation environment, testing different actions and observing the resulting states and rewards.

\textbf{Environment.}
A reliable simulation environment is essential, as it allows the agent to perform a large number of simulated runs (i.e., rollouts) and to explore the environment by trying out different actions. In our case, the environment is provided by the Grid2Op framework developed by RTE and simulates the load flow dynamics of a power grid.

\textbf{Action Space.}
The action space defines the kind of actions the agent can take to interact with the environment. In the general power grid setting, the agent can perform redispatching/curtailment actions as well as bus and line splitting (topology) actions. Due to the large number of possible topology configurations, the complexity of the topological action space poses a challenge for both human and RL-based agents.

\textbf{Observation.}
The current state of the environment, as observed by the agent. 
In the power grid setting this includes, among others, the  line loads, topology configuration as well as the  state of generator and loads.

\textbf{Reward.}
During training, the agent receives a reward~guiding it toward desired actions.
We use a reward based on cumulative line overflow, which the agent is trying to minimize.


\subsection{Grid2Op Environment}
\label{subsec:background-env}

Here, we described the Grid2Op~\cite{donnot2020grid2op} environment developed by RTE that we use for simulation.

\textbf{Components and Objectives.}
The Grid2Op environment models the standard components found in real-world power grids, including generators of different types, loads, storages, interconnections to other grids, and substations and lines connecting the grid. Experiments are carried out on the WCCI 2022 power grid (provided with the Grid2Op package) consisting of 118 substations, 186 power lines, 62 generators, 91 loads and 7 storages. The primary objective is to keep the grid in operation, avoiding congestions that can lead to automatic disconnections of overloaded lines, potentially isolating some of the generators and loads, resulting in a diverging power flow (i.e., an emergency state, potentially leading to black out).

\textbf{Chronics.}
Each environment comes with a set of scenarios, or chronics, that specify how power consumption (= loads) and production (= generators, mainly renewables which depend on external factors like weather) fluctuate over time. We utilize chronics provided by RTE as part of the 2022 L2RPN competition environment.

\textbf{Simulation.}
The agent has the capability of testing the effect of a chosen action by utilizing a limited simulation functionality based on a forecast of the chronics.

\textbf{Actions.}
Every 5 minutes, the agent can issue a combination of a continuous redispatch/curtailment action, a discrete topology action and continuous storage manipulation action. The topology action changes the bus configuration of a line or substation in the grid. The redispatch action must be in line with maximum allowed ramp ups/downs.

\subsection{Congestion Management Agent}
\label{subsec:background-agent}

This section summarizes the main components of our congestion management agent: the learned topology agent, the redispatching controller and the auxiliary modules building on top of it. For more details, please refer to~\cite{dorfer2022topo}.

\textbf{Topology Optimization Agent.}
The topology optimization agent is trained using a modified version of AlphaZero~\cite{silver2017alphazero}, where the agent interacts with the environment with the help of Monte Carlo Tree Search (MCTS). In each training iteration, the agent performs a number of rollouts, building a search tree in every environment step to select the best action. The trajectories are collected and used to update the policy for the next iteration, focusing on high-value states and actions. In deployment, the agent can be run with or without tree search.

\textbf{Redispatching Optimizer.}
A redispatching optimizer is deployed together with the RL-based topology optimization agent. When a congestion occurs, the remediation measure (redispatching or topology optimization) that reduces the maximum line load further (based on the simulation) is employed. In cases when neither of these two relieve the congestion, a combined redispatching and topology action is issued, selected from the top-5 topology action candidates, each with its own optimized redispatch.

\textbf{Redispatching/Topology Optimization Preference.}
Building on top of the topology optimization agent and redispatching optimizer published in \cite{dorfer2022topo}, we add a preference parameter $\alpha \in [0, 1]$. This enables the operator to set if any action type should be preferred over the other. With this parameter, grid operators can, for example, set redispatching preference, instructing the agent to pick a topology action only if the benefits of it over redispatching are substantial.

\textbf{Safe-State Skipping.}
In most time steps, the grid is running safely without congestions, and launching the agent is not necessary. In such cases, either a recovery (see below) or a no-operation action is performed.

\textbf{Topology Recovery.}
In safe states, if the grid is not in the default topology (all connections on bus 1, no bus splitting taking place), an automated default topology recovery takes place (this can be a multi-step process).

\textbf{Redispatch Reset.}
In safe states, if redispatching and curtailment are present in certain generators,
an automated redispatch reset takes place trying to reset generators to the planned dispatch
(this can be a multi-step process).

\subsection{Previous Results}
\label{subsec:background-previous-results}

\textbf{Results in prior work.}
In prior work \cite{dorfer2022topo}, the RL-based topology optimization agent combined with the redispatching optimizer has shown to outperform the redispatching-only approach both in survival and in operational costs, showing survival rate of 82\% (compared to 75\% with redispatching only) and decreasing the redispatching costs by 60\% on the L2RPN WCCI 2022 competition power grid.

\textbf{From simulation to real world.}
We expect that in the real world, the savings might be lower---both due to differences between real and simulated grids, as well as due to other factors, such as grid operators avoiding some of the less impactful topology actions in favor of redispatching to reduce the number of switching operations. However, due to the high costs of redispatching, even significantly lower savings would be very impactful. Hence, we do believe this technology has a great potential for a real world setting.

\section{AI-Assisted Day-Ahead Operational Planning}
\label{sec:day_ahead_planning}

The maintenance and operational planning of power grids is usually done in multiple stages corresponding to increasingly narrow time-scales ranging from full year ahead plans up to real time response where plans are continuously refined as more accurate data becomes available. For all planning stages, forecasts for demand and generation as well as the introduction of new grid components and expected outages (maintenance) are the basis for viable operational plans \footnote{For in-depth information refer to Commission Regulation (EU) on establishing a guideline on electricity transmission system operation \url{https://eur-lex.europa.eu/legal-content/EN/TXT/?uri=CELEX:02017R1485-20210315}}. 

During day-ahead planning, forecasting data for the next day are already available. Combined with an accurate power grid simulation, it provides the required environment for the deployment of the RL-based agent, as well as for evaluating the resulting operation plans.

Using the trained topology policy obtained from the training regime explained in Section~\ref{sec:background}, we are able to deploy our policy on previously unseen scenarios. In each state the \emph{Grid State Observer} judges if the current grid-state is unsafe and an action is required. If this is the case, the agent can choose between three options: the plain topology action guided by the trained policy, a plain redispatching action (based on a load flow calculation) and a combination of both topology and redispatching if one alone does not relieve the congestion based on the simulation. The preference of the agent towards topology vs. redispatching actions is controlled with a parameter $\alpha \in [0,1]$ in order to customize the resulting operational plan to the preference of the operator and the constraints of the grid. The tradeoff between the number of switching operations (and consequently the distance to the reference topology) and the effective redispatching cost can be controlled with this parameter, where $\alpha=0.0$ corresponds to plain redispatching actions and $\alpha=1.0$ to no redispatching at all. As a result we are able to produce multiple operational plans for a single scenario
and compare them based on the following metrics:

\begin{itemize}
    \item The number of \emph{switching operations} is the sum of individual bus switching operations at substations over the course of the day.
    \item The \emph{remaining congestion} is the sum of overloads of each line over the timespan of the day in MWh.
    \item The \emph{redispatch} applied to balance the grid in MWh.
\end{itemize}

\begin{table}[t]
  \centering
  \begin{tabular}{lrrr}
    \toprule
    Agent/$\alpha$    & Remaining &  Switching & Redispatch \\
    &Congestions&Operations& \\
    \midrule
    NoOp (BL) & $100.00 \% $& $0.00 \% $& $0.00 \% $\\
    $\alpha=0.0$ & $0.68 \% $& $0.00 \% $& $100.00 \% $\\
    $\alpha=0.25$ & $0.48 \% $& $2.82 \% $& $95.71 \% $\\
    $\alpha=0.5$ & $0.39 \% $& $56.59 \% $& $20.38 \% $\\
    $\alpha=0.75$ & $0.37 \% $& $84.61 \% $& $3.35 \% $\\
    $\alpha=1.0$ & $5.20 \% $& $100.00 \% $& $0.00 \% $\\
    \bottomrule
  \end{tabular}
  \caption{Comparison of our agent with different values for the parameter $\alpha$ on the 245 individual days with congestions of the WCCI 2022 evaluation episodes. All measures are averaged by overall steps to allow for a fair comparison before normalizing them by the maximum value of the corresponding column; \textit{remaining congestions} is normalized by the value for the NoOp Policy, \textit{switching operations} is normalized by the result from the topology-only configuration; \textit{redispatch} is normalized by the value of the redispatching-only configuration.}
\label{tab:exp-results}
\end{table}

\subsection{Experimental Evaluation}
\label{sec:sensitivity_ analysis}

To evaluate the performance of our agent, we test it on a held out dataset of the WCCI 2022 grid. The data consists of 52 scenarios, each spanning the course of one week, corresponding to the 52 weeks of the exemplary year 2050. To test day-ahead planning capabilities in a realistic setting, we split each week into seven days (364 days in total), disable adversarial attacks and remove the 118 days where no congestion occurs. The results of running our agent with five different settings of $\alpha$ on the remaining 246 day-long scenarios are summarized in Table~\ref{tab:exp-results} and compared to the NoOp baseline obtained by performing no operations on the grid at all.
Experimental results show how the amount of required redispatching decreases with increasing topological preference $\alpha$ while on the other hand the number of switching operations increases.
As not all congestions can be alleviated by the agents, all 'mixed' configurations ($\alpha \in \{0.25, 0.5, 0.75\}$) outperform redispatching-only or topology-only plans with respect to congestion reduction
while at the same time reducing redispatching compared to the redispatching only setting.
We also observe that not all congestions can be relieved with topology optimization alone ($\alpha=1.0$).

Although the amount of redispatching as well as the number of switching operations do not scale linearly with $\alpha$, we still believe that our results demonstrate how such a preference parameter can be utilized to trade-off redispatching costs with switching operations. The preferable configuration of the parameter might even vary from situation to situation depending on the type of congestion, day of the year, power grid and power grid operator preference. Consequently, having a multitude of diverse operational plan candidates available should be of great value when planing the operation of the power grid for the coming day.

\begin{figure}[t]
    \begin{subfigure}{.99\columnwidth}
      \caption{Congestion Profile}
      \includegraphics[width=.99\columnwidth]{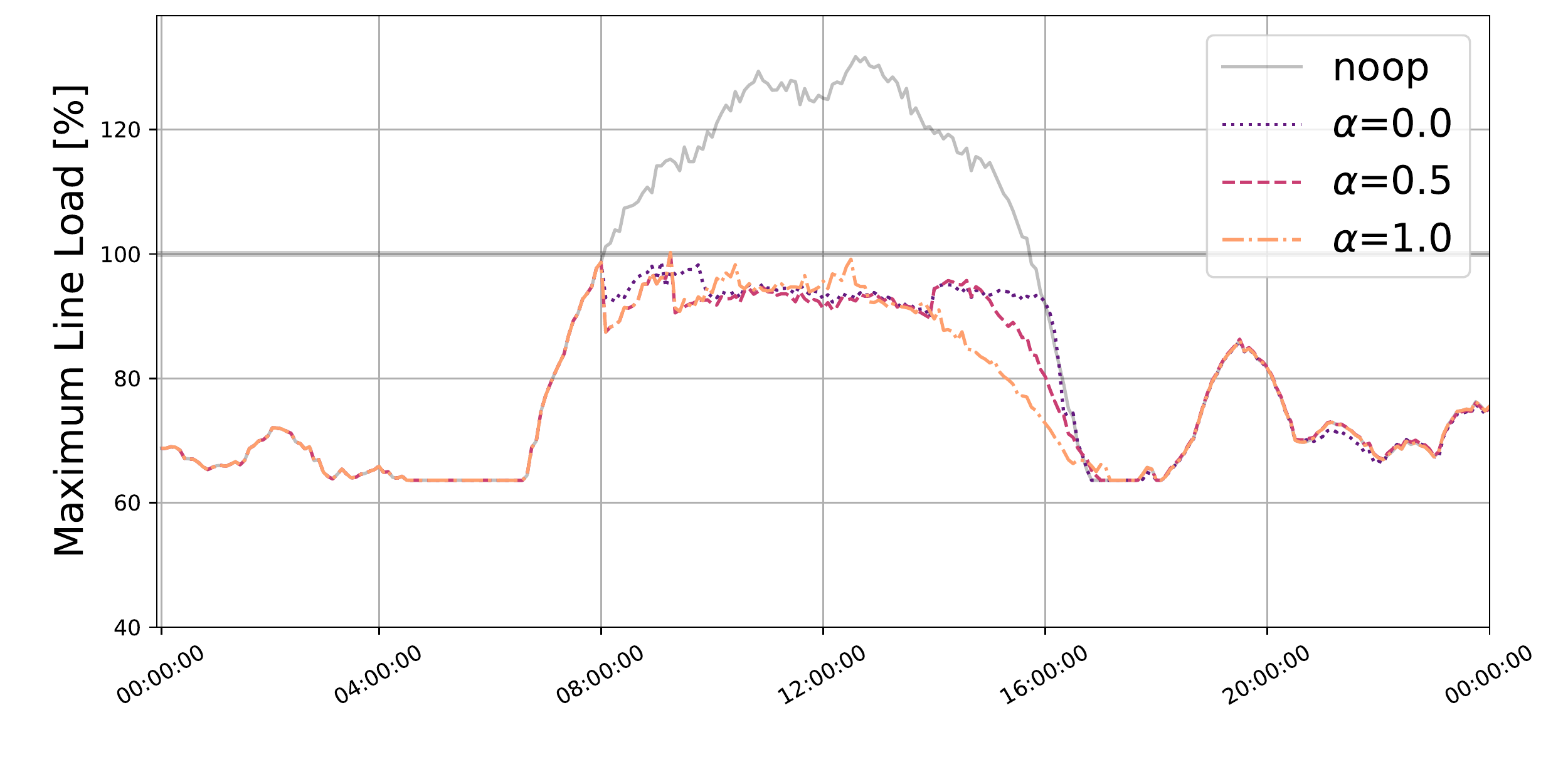}
      \label{fig:congestion_profile}
    \end{subfigure}
    \begin{subfigure}{.99\columnwidth}
      \caption{Redispatching Profile}
      \includegraphics[width=.99\columnwidth]{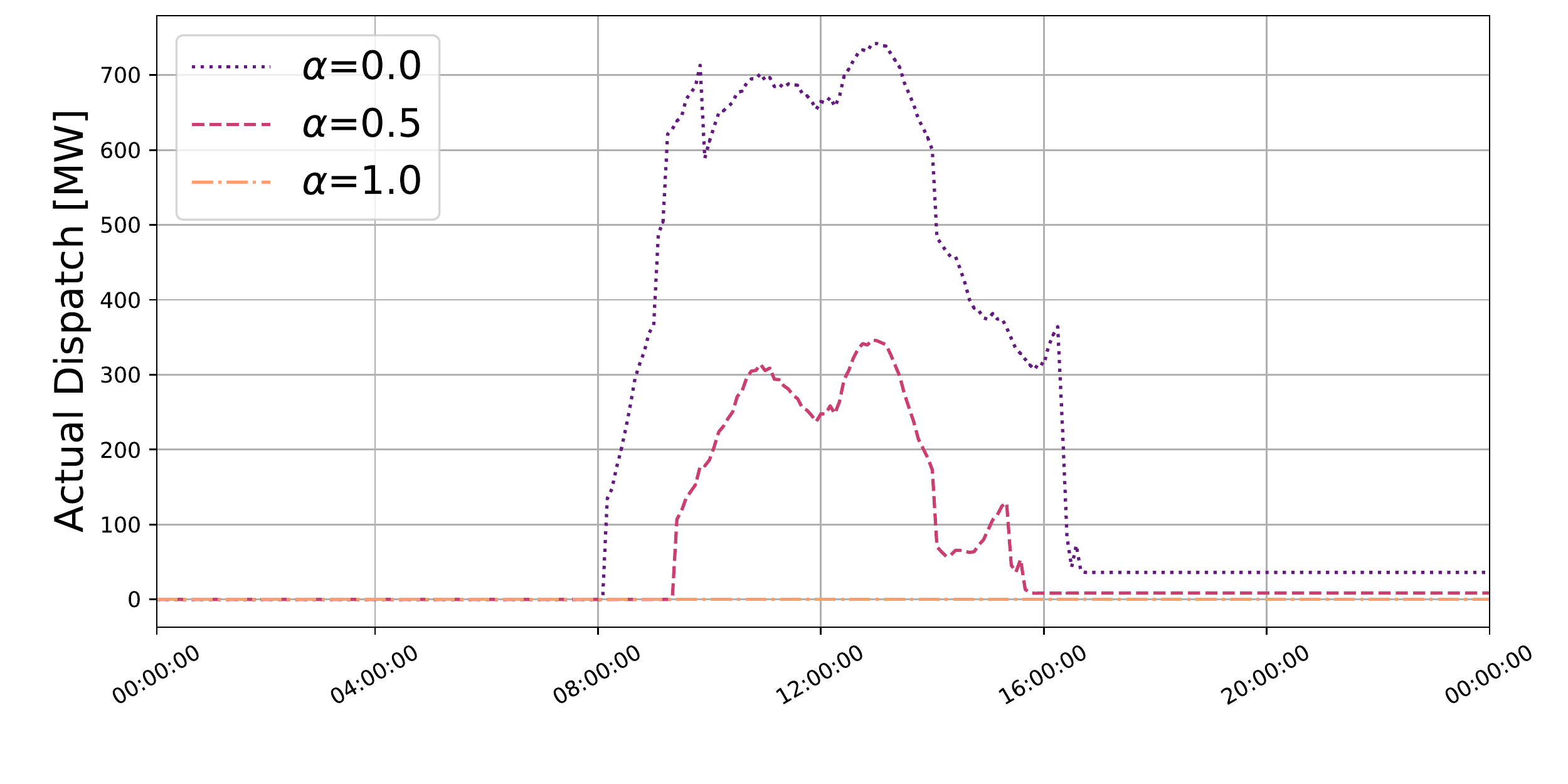}
      \label{fig:redispatching}
    \end{subfigure}
    \begin{subfigure}{.99\columnwidth}
      \caption{Distance to reference topology}
      \includegraphics[width=.99\columnwidth]{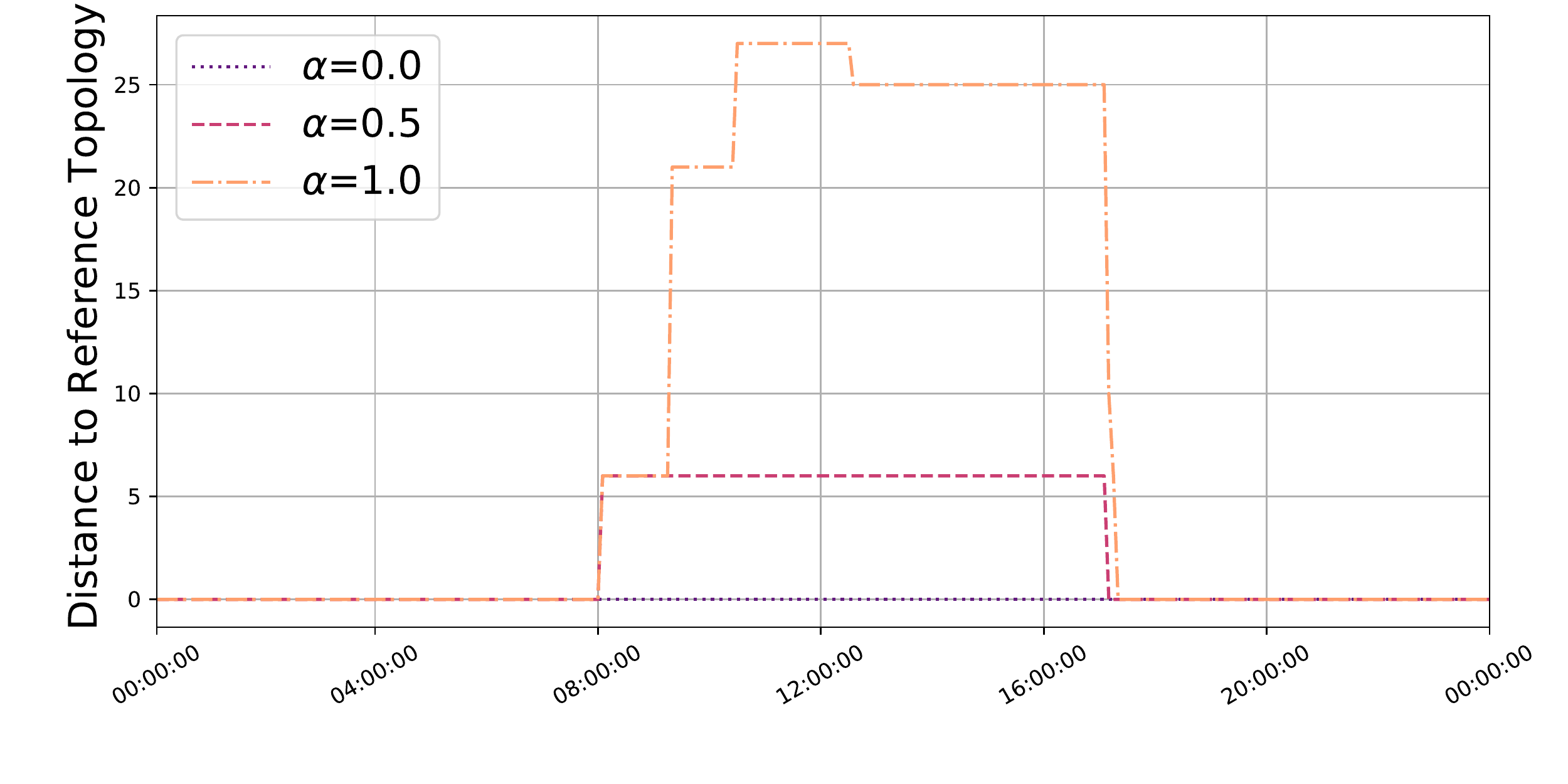}
      \label{fig:distance_to_reference_topology}
    \end{subfigure}
    \caption{Comparison of three different agents on one example scenario covering a planning period of 24 hours. Here the distance to reference topology refers to the sum of line end points that need to be switched to a different bus to return to the reference topology.}
    \label{fig:example_day_comparison}
\end{figure}
Figure~\ref{fig:example_day_comparison} gives a more nuanced insight into the behaviour of the different agent configurations for one particularly challenging congestion management scenario on a summer day in the year 2050. First, note that each of the three agents is able to solve the congestion by deploying completely different strategies. Most interesting however is the behaviour of the mixed agent (with $\alpha = 0.5$). When the maximum line load is first expected to surpass the $100\%$ utilization threshold at 8:00am it deploys a switching operation to alleviate the congestion similar to the topology only policy. Shortly after ($\sim$ 9:30am) as the maximum line load of the NoOp policy starts approaching $120\%$ additional congestion management is required. While the $\alpha=1.0$ policy continues altering the topology of the grid the mixed agent elevates the remaining congestion with a moderate redispatching action. As a result it saves about $60\%$ of the redispatching compared to the redispatching-only policy and about $80\%$ of the bus switching operations compared to the topology-only policy.

\subsection{Sensitivity Analysis}
\label{sec:sensitivity_ analysis}
In an attempt to facilitate further understanding of robustness of the proposed topology actions, we conduct an elementary sensitivity analysis. In this setting, we create topology-only operational plans for the original evaluation chronics, and then evaluate them (i.e., compare to the no-operation policy in this case) on modified chronics that represent an inaccurately forecasted future. In this simplistic setting, we have four modified chronics: decreasing and increasing the outputs of all wind generators by 10\%, and then imbalancing the grid by increasing the wind power generation in the eastern part by 25\% while decreasing it by 25\% in the western part, and vice versa. 
We choose wind because in the WCCI 2022 evaluation chronics wind already fluctuates significantly throughout the scenarios while at the same time contributes a significant portion of overall power (around a quarter); for the imbalanced scenarios, we split the grid into western and eastern parts as this is a natural split into two more strongly connected components considering the grid topology\footnote{Refer to the challenge introduction for the energy mix and grid topology overview: \url{https://codalab.lisn.upsaclay.fr/competitions/5410}}.
While these are just four arbitrary scenarios, we still think they facilitate a basic understanding of how topology actions behave in less predictable cases. In the real world, such scenarios would represent alternative forecasted futures, and should help better evaluate the robustness of considered congestion remediation measures (in fact, we are already considering addition of a more elaborate sensitivity module into the AI assistant presented in Section \ref{sec:ai_assisted_control}).

\begin{figure}[h]
  \includegraphics[width=\columnwidth]{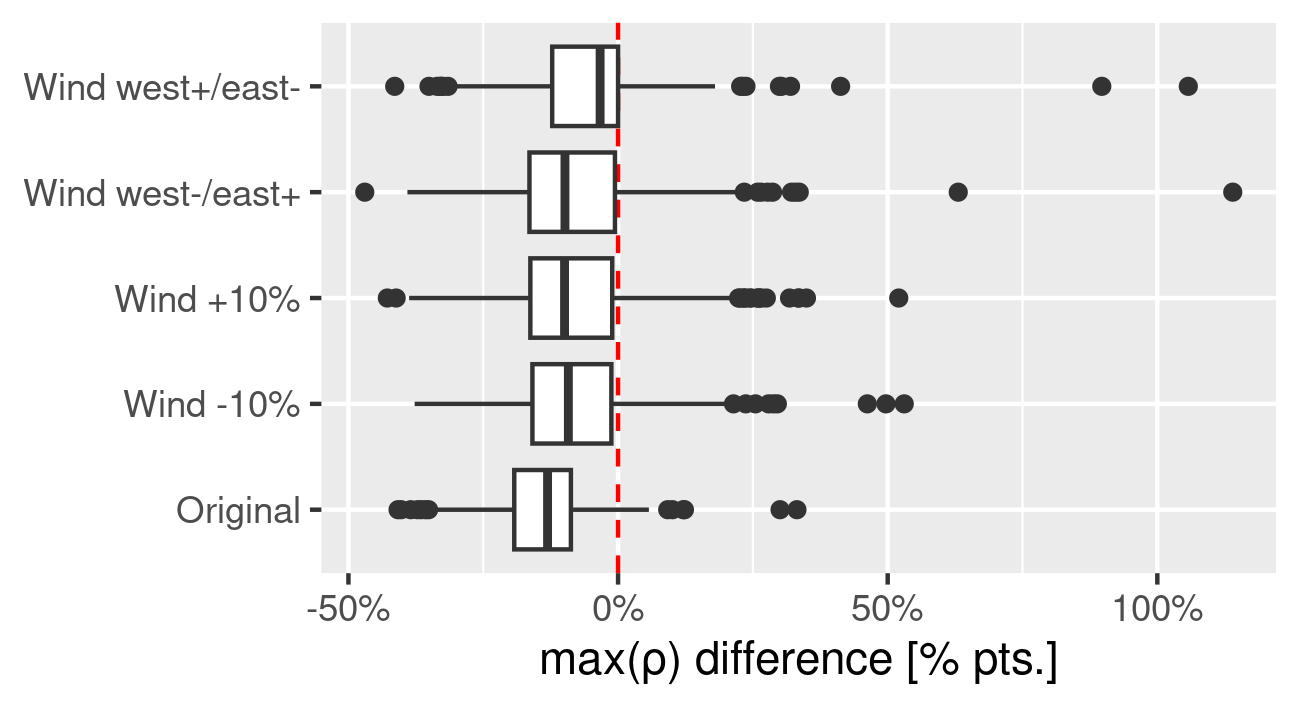}
  \caption{Sensitivity analysis. The evaluation chronics are split into individual congestions (i.e., consecutive overflows). Maximum encountered line load $\max(\rho)$ is calculated per congestion, comparing the topology plan to the no-operation policy, plotting the differences. Difference below 0\% pts. means the topology plan relieved congestion better than the no-op.}
  \label{fig:sensitivity_analysis_boxplot}
\end{figure}
As seen in Figure \ref{fig:sensitivity_analysis_boxplot},
the majority of the proposed topology operations are reasonably robust against the tested scenarios, with over 75\% of actions still reducing congestion compared to the no-operation policy, even when the chronics develop differently to what the topology actions were designed for. This does not mean that congestions were completely resolved in all of these cases (some additional ad-hoc redispatching might still be required), but it shows that in most cases, topology actions still helped the stabilize the grid. 

However, as shown by the box plot whiskers and individual outliers on the right, there is also a minor share of topology actions that performed significantly worse than the no-operation on the modified chronics, leading to potentially disastrous overflows. While a few such outliers are present even on the original chronics, representing scenarios that the topology-only agent failed to treat well (remember we are evaluating the topology-only agent here, without any additional redispatch), there are many more such outliers in the modified scenarios, often with significantly larger magnitude.

Hence, results of this elementary sensitivity analysis suggest that while the majority of topology operations seem to be reasonably stable against significantly imprecise predictions, there are some more volatile topology actions among them which should be avoided. These findings imply that grid operators might benefit from integration of a similar robustness analysis into tools such as the AI-assisted control workflow described in Section \ref{sec:ai_assisted_control}, emphasizing the importance of the projection component suggested in \cite{prostejovsky2019future}.

\section{AI-Assisted Control}
\label{sec:ai_assisted_control}


%
\begin{figure*}[t]
  \centerline{\includegraphics[width=0.97\textwidth]{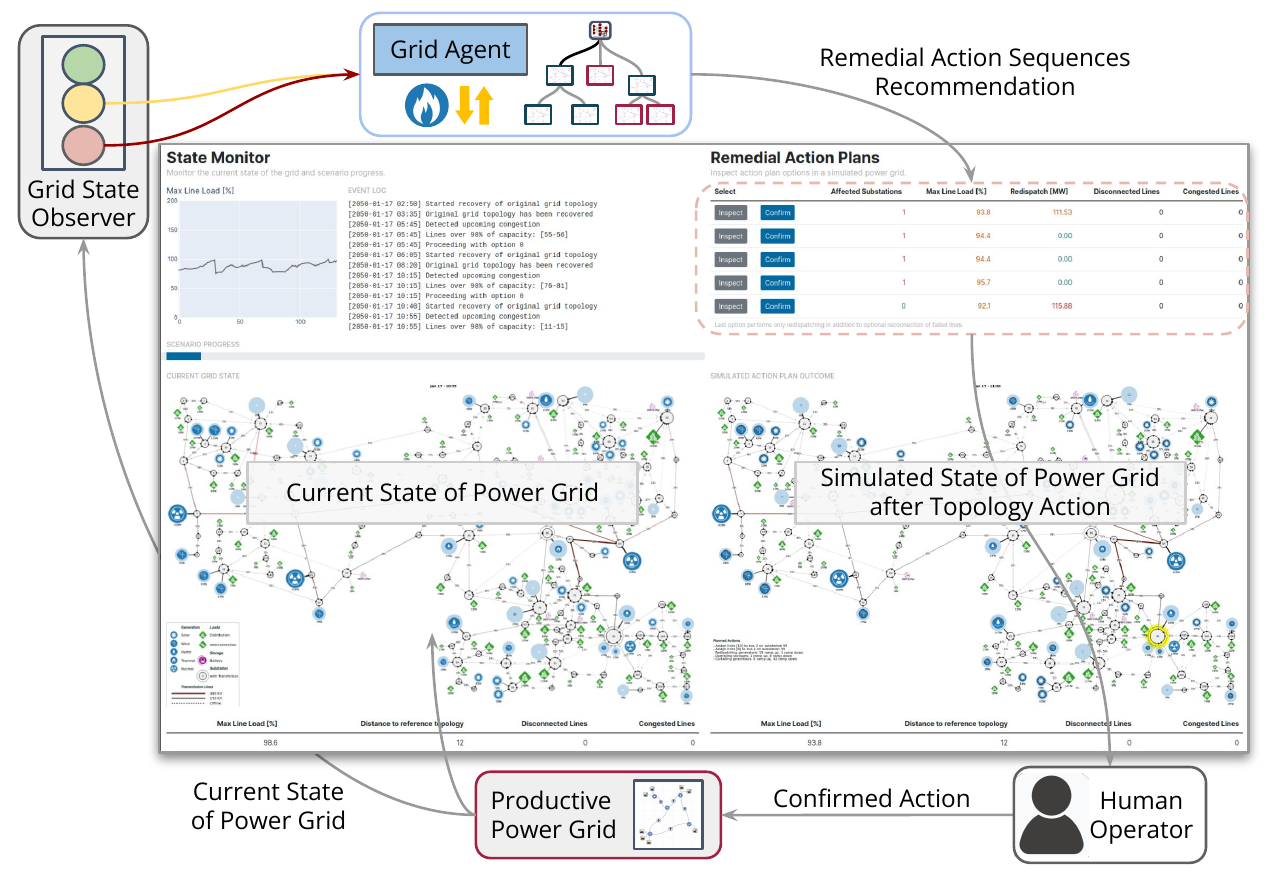}}
  \caption{Overview of RL-agent-powered real-time remedial action recommendation assistant (\url{http://grid-demo.enlite.ai/}).}
  \label{fig:assistant_overview}
\end{figure*}
This section introduces AI-assisted control
as a second application area for the power grid AI agent proposed in \cite{dorfer2022topo} and summarized in Section \ref{subsec:background-agent}.
In particular, we describe how to embed and utilize an agent,
originally trained for fully autonomous control,
within a human-in-the-loop decision support system.
The central design principle of the proposed AI assistant is
to support and enhance human decision making
by recommending viable action scenarios along with augmenting information,
projecting and explaining how suggested operational recommendations will most likely impact the live system.
However, and most importantly, we leave the final decision to the human operator
to preserve human control and accountability.
These design choices are motivated and in line with prior work and studies
on trustworthy and interpretable AI systems in the context of powergrids~\cite{prostejovsky2019future,marott2022towards,viebahn2022potential}.
The workflow of the AI assistant is summarized in Figure~\ref{fig:assistant_overview}
and covers the following steps:
\begin{itemize}
    \item A \emph{Grid State Observer} continuously monitors the current state of the live power grid with respect to operational limits and unexpected events (e.g., line outages).
    \item Once a non-safe state is encountered (e.g., a power line is expected to surpass its thermal limits) the agent starts a policy-guided tree search to discover a set of viable action scenarios (e.g., a single or even a sequence of grid operation actions) that are capable of recovering from this critical situation (e.g., relieving the congestion).
    \item A ranked list of action candidates -- the top results of the tree search -- is presented to the human operator in a graphical user interface for evaluation (testing the impact of an action candidate in a load flow simulation) and selection. Along with the potential for relieving the congestion other grid specific safety considerations are taken into account (e.g., a contingency analysis for n-1 stability of the respective resulting states). 
    \item The human operator evaluates the provided set of suggested topology change actions.
    \item Once satisfied, the operator confirms the best action candidate for execution on the live power grid.
    \item The selected action candidate is applied to the power grid and the resulting state is again fed into the \emph{Grid State Observer} and visualized for human operators in the graphical user interface. This step closes the human-in-the-loop workflow.
\end{itemize}

When operating a real power grid, most congestion is anticipated and addressed in advance to ensure safe operating conditions at all times (see Section~\ref{sec:day_ahead_planning}). The time window in which a remediation measure has to be implemented to prevent the grid from becoming unstable varies from several hours to a few minutes in the event of unexpected fluctuations or equipment failure. Our AI assistant demonstrates that recommendations can be produced in less than three second, meaning human operators can incorporate the agent’s outputs into their evaluation procedures without loosing valuable time for critical decisions.

\section{Conclusion}
\label{sec:discussion}
As part of this paper we combine a recently proposed RL-based topology agent~\cite{dorfer2022topo} with a configurable portion of redispatching and
show how to utilize such an agent to assist the two real-world applications, AI-assisted \textit{day-ahead planning} and \textit{real-time control}.
Our experimental evaluations reveal that the proposed system is able to produce multiple viable operational plans where the amount of switching operations vs. redispatching actions can be controlled with a single parameter, to the operator's preference.
Furthermore, in our simulated setting, all combined approaches outperform a redispatching-only baseline, reducing the remaining congestions further while also decreasing the redispatching cost by 75\% (with $\alpha = 0.5$) compared to the redispatching-only configuration.
Additionally, we propose and implement a concrete instance of an AI-assisted real-time remedial action suggestion demonstrator.
The assistant utilizes state-of-the-art machine learning methods
to assist human operators with viable real-time action recommendations.
Along with the action candidates it presents augmenting information
on the projected impact of these candidates on the live grid
via forecasting and simulation.
We hope that our work helps to bring this promising technology closer to the control room, contributing to a safe and sustainable operation of increasingly complex power grids of tomorrow.

{
\bibliographystyle{abbrv}
\bibliography{main}

\begin{thebibliography}{10}

\bibitem{donnot2020grid2op}
B.~Donnot.
\newblock {Grid2op: A testbed platform to model sequential decision making in
  power systems.}
\newblock \url{https://GitHub.com/rte-france/grid2op}, 2020.

\bibitem{dorfer2022topo}
M.~Dorfer, A.~R. Fuxj{\"{a}}ger, K.~Koz{\'{a}}k, P.~M. Blies, and M.~Wasserer.
\newblock Power grid congestion management via topology optimization with
  alphazero.
\newblock {\em RL4RealLife Workshop in the 36th Conference on Neural
  Information Processing Systems (NeurIPS 2022)}, 2022.

\bibitem{luo2022cooling}
J.~Luo, C.~Paduraru, O.~Voicu, Y.~Chervonyi, S.~Munns, J.~Li, C.~Qian,
  P.~Dutta, J.~Q. Davis, N.~Wu, et~al.
\newblock Controlling commercial cooling systems using reinforcement learning.
\newblock {\em RL4RealLife Workshop in the 36th Conference on Neural
  Information Processing Systems (NeurIPS 2022)}, 2022.

\bibitem{marot2021l2rpntrust}
A.~Marot, B.~Donnot, K.~Chaouache, A.~Kelly, Q.~Huang, R.~Hossain, and J.~L.
  Cremer.
\newblock Learning to run a power network with trust.
\newblock {\em CoRR}, abs/2110.12908, 2021.

\bibitem{marot2021retrospective}
A.~Marot, B.~Donnot, G.~Dulac{-}Arnold, A.~Kelly, A.~O'Sullivan, J.~Viebahn,
  M.~Awad, I.~Guyon, P.~Panciatici, and C.~Romero.
\newblock Learning to run a power network challenge: a retrospective analysis.
\newblock {\em CoRR}, abs/2103.03104, 2021.

\bibitem{marot2019l2rpn}
A.~Marot, B.~Donnot, C.~Romero, L.~Veyrin{-}Forrer, M.~Lerousseau, B.~Donon,
  and I.~Guyon.
\newblock Learning to run a power network challenge for training topology
  controllers.
\newblock {\em CoRR}, abs/1912.04211, 2019.

\bibitem{marot2022perspectives}
A.~Marot, A.~Kelly, M.~Naglic, V.~Barbesant, J.~Cremer, A.~Stefanov, and
  J.~Viebahn.
\newblock Perspectives on future power system control centers for energy
  transition.
\newblock {\em Journal of Modern Power Systems and Clean Energy},
  10(2):328--344, 2022.

\bibitem{marott2022towards}
A.~Marot, A.~Rozier, M.~Dussartte, L.~Crochepierre, and B.~Donnot.
\newblock Towards an ai assistant for power grid operators.
\newblock {\em arXiv}, 2020.

\bibitem{osipychev2022airtraffic}
D.~Osipychev and D.~Margineantu.
\newblock Reinforcement learning-based air traffic deconfliction.
\newblock {\em RL4RealLife Workshop in the 36th Conference on Neural
  Information Processing Systems (NeurIPS 2022)}, 2022.

\bibitem{prostejovsky2019future}
A.~M. Prostejovsky, C.~Brosinsky, K.~Heussen, D.~Westermann, J.~Kreusel, and
  M.~Marinelli.
\newblock The future role of human operators in highly automated electric power
  systems.
\newblock {\em Electric Power Systems Research}, 175:105883, 2019.

\bibitem{schrittwieser2020muzero}
J.~Schrittwieser, I.~Antonoglou, T.~Hubert, K.~Simonyan, L.~Sifre, S.~Schmitt,
  A.~Guez, E.~Lockhart, D.~Hassabis, T.~Graepel, et~al.
\newblock Mastering atari, go, chess and shogi by planning with a learned
  model.
\newblock {\em Nature}, 588(7839):604--609, 2020.

\bibitem{silver2017alphazero}
D.~Silver, T.~Hubert, J.~Schrittwieser, I.~Antonoglou, M.~Lai, A.~Guez,
  M.~Lanctot, L.~Sifre, D.~Kumaran, T.~Graepel, et~al.
\newblock Mastering chess and shogi by self-play with a general reinforcement
  learning algorithm.
\newblock {\em arXiv preprint arXiv:1712.01815}, 2017.

\bibitem{stott1974loadflow}
B.~Stott.
\newblock Review of load-flow calculation methods.
\newblock {\em Proceedings of the IEEE}, 62(7):916--929, 1974.

\bibitem{viebahn2022potential}
J.~Viebahn, M.~Naglic, A.~Marot, B.~Donnot, and S.~H. Tindemans.
\newblock Potential and challenges of ai-powered decision support for
  short-term system operations.
\newblock {\em CIGRE Session 2022}, 2022.

\end{thebibliography}
}

\end{document}